\documentclass{article}
\usepackage[utf8]{inputenc}
\usepackage{iclr2020_conference,times}
\usepackage{todonotes}
\usepackage{amsmath,mleftright}
\usepackage{multirow}
\usepackage{todonotes}
\usepackage{wrapfig}
\usepackage{subcaption}
\title{Meta-Learning and Self-Supervised Pretraining for Real World Image Translation}

\author{Ileana Rugina\thanks{Equal contribution.} \\
MIT EECS \\
\texttt{irugina@mit.edu}
\And
Rumen Dangovski\textsuperscript{$*$} \\
MIT EECS \\
\texttt{rumenrd@mit.edu}
\And
Mark Veillette \\
MIT Lincoln Lab \\
\texttt{mark.veillette@ll.mit.edu}
\AND
Pooya Khorrami \\
MIT Lincoln Lab \\
\texttt{pooya.khorrami@ll.mit.edu} 
\And
Brian Cheung \\
MIT CSAIL \& BCS \\
\texttt{cheungb@mit.edu} 
\And
Olga Simek \\
MIT Lincoln Lab \\
\texttt{osimek@ll.mit.edu} 
\AND
Marin Solja\v{c}i\'{c} \\
MIT Physics\\
\texttt{soljacic@mit.edu}
}

\usepackage{hyperref}
\hypersetup{
    colorlinks,
    citecolor=orange,
    filecolor=black,
    linkcolor=black,
    urlcolor=black
}

\usepackage[]{algorithm2e}
\RestyleAlgo{boxruled}

\iclrfinalcopy

\usepackage{amsmath,amsfonts,bm}

\def\eqref#1{equation~\ref{#1}}

\def\1{\bm{1}}

\DeclareMathAlphabet{\mathsfit}{\encodingdefault}{\sfdefault}{m}{sl}
\SetMathAlphabet{\mathsfit}{bold}{\encodingdefault}{\sfdefault}{bx}{n}

\begin{document}

\maketitle

\begin{abstract}
Recent advances in  deep learning, in particular enabled by hardware advances and big data, have provided impressive results across a wide range of computational problems such as computer vision, natural language, or reinforcement learning. Many of these improvements are however constrained to problems with large-scale curated data-sets which require a lot of human labor to gather. Additionally, these models tend to generalize poorly under both slight distributional shifts and low-data regimes. In recent years, emerging fields such as meta-learning or self-supervised learning have been closing the gap between proof-of-concept results and real-life applications of machine learning by extending deep-learning to the semi-supervised and few-shot domains.
We follow this line of work and explore spatio-temporal structure in a recently introduced image-to-image translation problem in order to:  \emph{i)} formulate a novel multi-task few-shot image generation benchmark and \emph{ii)} explore data augmentations in contrastive pre-training for image translation downstream tasks. We present several baselines for the few-shot problem and discuss trade-offs between different approaches. Our code is available at \url{https://github.com/irugina/meta-image-translation}.

\end{abstract}

\section{Introduction}
Benchmarks such as ImageNet \citep{imagenet_cvpr09imagenet_cvpr09} in computer vision or SQuAD \citep{rajpurkar2016squad} in natural language processing have been pivotal in popularizing deep-learning techniques and demonstrating their power. More recently, works such as ObjectNet \citep{NEURIPS2019_97af07a1} in vision have shown impressive performance on these established benchmarks does not translate to good performance in real-world situations, where the datasets might be less structured or more diverse. There is a lot of interest in devising more challenging datasets, both of general interest as well as domain-specific applications, that more closely resemble real-world situations practitioners might encounter when trying to deploy machine learning models.
Growing fields such as self-supervised \citep{Le_Khac_2020} or multi-task learning \citep{hospedales2020metalearning} reflect these interests and provide promising solutions to the aforementioned issues.

However, the problem of model evaluation remains: for example, in few-shot learning model evaluation is currently largely constrained to Omniglot \citep{lake2019omniglot, Lake2015HumanlevelCL} (which has essentially been saturated),  Miniimagenet \citep{vinyals2017matching} and Metadataset \citep{DBLP:journals/corr/abs-1903-03096}.
Similarly, contrastive pretraining techniques are generally evaluated on ImageNet.

We address these known limitations in our field by contributing a new computer vision multi-task problem and move away from classification problems towards the field of image-generation by leveraging a weather dataset \citep{NEURIPS2020_fa78a161} to formulate a novel few-shot image-to-image translation problem. In doing so we use the spatio-temporal structure of this dataset in construction of the few-shot tasks. We further leverage this structure through contrastive pretraining by exploring novel data augmentations and show consistent improvements in sample quality. Our work has three main contributions:
\begin{itemize}
    \item we introduce a novel few-shot image translation benchmark and provide several baselines for this problem.
    \item we train generative adversarial networks using model-agnostic meta-learning (MAML)~\citep{pmlr-v70-finn17a} and discuss the advantages and drawbacks of this approach.
    \item we pretrain part of the generator parameters using contrastive learning and show consistent improvements in downstream image-generation performance.
\end{itemize}

\section{Background and Related Work}

\subsection{Storm Event Imagery}
\begin{figure}[t]
    \centering    \includegraphics[width=1\linewidth]{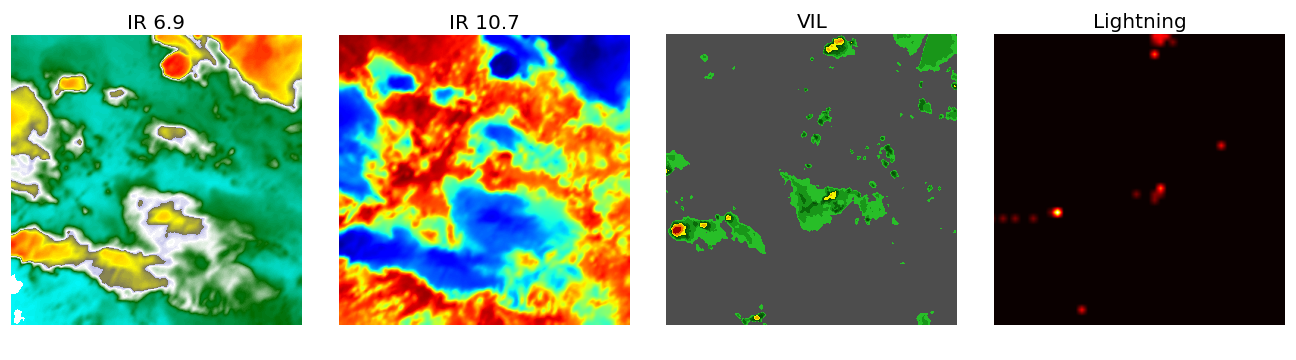}
    \caption{{\bf  Frame from The Storm Event Imagery (SEVIR) dataset.} We use four of the five available modalities: 2 IR, VIL, and lightning information. }
    \label{fig:sevir}
\end{figure}

The Storm Event Imagery (SEVIR) \citep{NEURIPS2020_fa78a161} is a  radar and satellite meteorology dataset. It is a collection of over 10,000 weather events, each of which tracks 5 sensor modalities within $384\text{ km} \times 384\text{ km}$ patches for 4 hours. The events are uniformly sampled so that there are 49 frames for each 4 hour period, and the 5 channels consist of:
    \emph{i)} 1 visible and 2 IR sensors from the GOES-16 advanced baseline \citep{ACloserLookattheABIontheGOESRSeries}
    \emph{ii)} vertically integrated liquid (VIL) from NEXTRAD
    \emph{iii)} lightning flashes from GOES-16 
.
Fig. \ref{fig:sevir} shows examples of the two IR and the VIL modalities. We disregard the visible channel because it often contains no information as visible radiation is easily occluded. 
\citep{NEURIPS2020_fa78a161} suggested several machine learning problems that can be studied on SEVIR and provided baselines for two of these: nowcasting and synthetic weather radar generation. 
In both cases they train U-Net models to predict VIL information and experiment with various loss functions. 

\subsubsection{Evaluation}
We review common evaluation metrics used in the satellite and radar literature to analyse artificially-generated VIL imagery. They all compare the target and generated samples after binarizing them with an arbitrarily threshold in $[0, 255]$ and look at counts in the associated confusion matrix. Let $H$ denote the number of true positives, $C$ denote the number of true negatives, $M$ denote the number of false negatives and $F$ the number of false positives. \citep{NEURIPS2020_fa78a161} define four evaluation metrics: Critical Succes Index {\bf (CSI)} is equivalent to the intersection over union $\frac{H}{H+M+F}$;  Probability of detection {\bf(POD)} is equivalent to recall $\frac{H}{H+M}$;  Succes Ratio {\bf(SUCR)} is equivalent to precision $\frac{H}{H+F}$.

\subsection{Generative Adversarial Networks}

There has been a lot of interest in training GANs in low-data settings. In this scenario the main challange is that the discriminator network can simply memorize the train-set and quickly reach perfect performance on known examples \citep{zhao2020differentiable}.  In this case training quickly becomes unstable and the generator is not able to create realistic samples. Additionally, the discriminator performs poorly when evaluated on held-out validation or test splits. 

The only prior work we are aware of on the topic of few-shot multi-task image generation with second-order gradient updates is that by \citet{DBLP:journals/corr/abs-1901-02199}. They optimize using Reptile \citep{nichol2018firstorder}, a first-order approximation to MAML, and evaluate on the  MNIST and Omniglot datasets. They also introduce a dataset which presents a very clear delimitation between different tasks and more generally does not exhibit the challenges of modeling real-world phenomena because the examples are icons rather than realistic images.

\citet{zhao2020differentiable} apply augmentations to both real and generated samples and require the transformations to be differentiable in order to backpropagate to the generator with good results using as little as $10\%$ of the available samples. 

Consistency Regularization(CR) is a semi-supervised training technique introduced to GANs by \citep{DBLP:journals/corr/abs-1910-12027} as a discriminator regularization method that can be used in conjunction with gradient normalization methods to encourages the discriminator's predictions to be invariant under arbitrarily transformations applied to real samples. 
\citet{zhao2020improved} extend this work to balanced Consistency Regularization (bCR) and latent Consistency Regularization (zCR), and combine the two into Improved Consistency Regularization (ICR). These techniques regularize the discriminator and generator network using data augmentations on the generated samples or latent variables.

\section{Few-Shot Benchmark and Our Baselines}
\label{sec:fsbench}

\subsection{Benchmark Construction}
We leverage the SEVIR \citep{NEURIPS2020_fa78a161} dataset to construct a few-shot multi-task image-to-image translation problem where each task corresponds to one event. From the $49$ available frames we keep the first $N_{\text{support}}$ frames to form the task's support set and the next $N_{\text{query}}$ to be the query. Throughout the following experiments we set $N_{\text{support}} = N_{\text{query}} = 10$. 

For the sake of this discussion let's assume we have re-scaled all input modalities to the maximum observed resolution  $384 \times 384$ so that we can view all of SEVIR as a simple input tensor $\mathcal{D}_1 \in \mathbb{R}^{N_{\text{event}} \times N_{\text{frames}} \times C \times w \times h}$, where: \emph{i)} $N_{\text{event}} = 11479$; \emph{ii)} $N_{\text{frames}} = N_{\text{support}} + N_{\text{query}}$; \emph{iii)} $C = 4$; \emph{iv)} $ w=h=384$. The four input channels are split into three input modalities $C_{\text{in}} = 3$ and one target $C_{\text{out}} = 1$. For joint training we ignore the hierarchical dataset structure and collapse the first two axis $\mathcal{D}_2 \in \mathbb{R}^{N \times C \times w \times h}$, where $N = N_{\text{event}} \times N_{\text{frames}}$ --- the total number of frames. 

\subsection{Methods}
We solve the aforementioned task using either first-order or second-order gradient descent methods on U-Nets trained using either reconstruction or adversarial objectives. Note that in the case when we train GANs using MAML we are searching for a good initialization for multiple related saddle-point problems. Despite this challenging task, we still obtain good performance. 

Below we present the meta-train loop for adversarial networks, which is a novel contribution of our work. 
For simplicity, we only present the variant with a single SGD inner-loop adaptation step.
We train a U-Net generator $G$ with model weights $w_G$ jointly with an extranous patch discriminator $D$ with model weights $w_D$  using data $\mathcal{D} \in  \mathbb{R}^{N_{\text{event}} \times N_{\text{frames}} \times C \times w \times h}$. We use batched alternating gradient descent as our optimization algorithm and consider batches $\mathcal{B} \in \mathbb{R}^{B \times N_{\text{frames}} \times C \times w \times h}$, where $B$ is the meta-batch size. Each of these can be split along the second axis into the support and query sets, and along the third axis into the source ($S$) and target tensors ($T$) to create 
$ S^{\text{support}} \in \mathbb{R}^{B \times N_{\text{support}} \times C_{\text{in}} \times w \times h}$, 
$S^{\text{query}} \in \mathbb{R}^{B \times N_{\text{query}} \times C_{\text{in}} \times w \times h}$,
$T^{\text{support}} \in \mathbb{R}^{B \times N_{\text{support}} \times C_{\text{out}} \times w \times h}$,
$T^{\text{query}} \in \mathbb{R}^{B \times N_{\text{query}} \times C_{\text{out}} \times w \times h}$. For any of these tensors $X \in \{ S^{\text{support}}, \; S^{\text{query}}, \; T^{\text{support}},  \; T^{\text{query}} \}$ we refer to the four-dimensional tensor given by the $i^{\text{th}}$ task or event as $X_i$. We use such four-dimensional tensor quantities to evaluate the generator and discriminator loss functions:
\begin{equation}
\label{eq:generator_loss}
    \hat{\mathcal{L}}_G (t^{\text{generated}}, t, s; w_G, w_D) = - \log{D(s, t^{\text{generated}})} + \lambda || t^{\text{generated}} - t||_1
\end{equation}
\begin{equation}
\label{eq:discriminator_loss}
   \hat{\mathcal{L}}_D (t^{\text{generated}}, t, s; w_G, w_D) = \frac{1}{2}\left( \log{D(s, t^{\text{generated}})}  - \log{D(s, t)}  \right),
\end{equation}
where $t^{\text{generated}} = G(s)$ is a generated target sample, $t$ and $s$ are corresponding input and output modalities, $||x||_1$ is the mean absolute error. Note the slight abuse of notation where by $\log D(x,y)$ with $x,y\in\mathbb{R}^{N\times C \times w \times h}$ we mean the average $\frac{1}{N} \sum_{i=1}^N \log{D(x_{i}, y_{i})}$. This formulation also uses the trick of replacing $\max \log{\left(1-D(G(z))\right)}$ with $\min \log{D(G(z))}$ to obtain a non-saturating generator objective. We wrote the loss functions above such that both players want to minimize their respective objectives. 

For each task in a meta-batch size we evaluate the losses above on the support set frames and adapt to this event using SGD to obtain parameters $\phi$. We then evaluate the same losses on the task's query set using finetuned models. We repeat these two steps for each event in the meta-batch and perform a second-order gradient update to the initial parameters to optimize the average loss across all events in the meta-batch. This procedure is schematically summarized in Algorithm~\ref{algo:maml_gan}. The procedure for the Reconstruction loss requires minimal modification from Algorithm~\ref{algo:maml_gan}. In particular, we remove every line related to the discriminator $D$ and modify Equation~\ref{eq:generator_loss} by removing the first term for the discriminator.

\begin{algorithm}[t]
 \For{\normalfont meta-train-batch $\mathcal{B} \in \mathbb{R}^{B \times N_{\text{frames}} \times C \times w \times h}$ }{
 
  unpack $\mathcal{B} \in \mathbb{R}^{B \times N_{\text{frames}} \times C \times w \times h}$ along support/query, source/target into: \\
  \hphantom{~~~~~~~~~~~~~~~~~~~~~~~~~~} $ S^{\text{support}}, \; S^{\text{query}}, \; T^{\text{support}},  \; T^{\text{query}}$ \\
  {\bf init} $l_G^{\text{batch}} = 0$, $l_D^{\text{batch}} = 0$\\
  \For{\normalfont each event $i$ out of $B$ in meta-batch} {
    forward pass $T_{i}^{\text{support; generated}} = G(S_i^{\text{support}})$ \\
    $l_G^{\text{adapt}} = \mathcal{L}_G(T_{i}^{\text{support; generated}}, T_{i}^{\text{support}},  S^{\text{support}}_i; w_G, w_D)$ from Eq.~\ref{eq:generator_loss} \\
    $l_D^{\text{adapt}} = \mathcal{L}_D(T_{i}^{\text{support; generated}}, T_{i}^{\text{support}},  S^{\text{support}}_i; w_G, w_D)$ from Eq.~\ref{eq:discriminator_loss}\\
    task-specific parameters $\phi_G \leftarrow w_G - \eta \nabla_{w_G} l_G^{\text{adapt}}$ \\
    task-specific parameters $\phi_D \leftarrow w_D - \eta \nabla_{w_D} l_D^{\text{adapt}}$ \\
    forward pass $T_{i}^{\text{query; generated}} = G(S_i^{\text{query}})$ \\
    $l_G = \mathcal{L}_G(T_{i}^{\text{query; generated}}, T_{i}^{\text{query}},  S^{\text{query}}_i; \phi_G, \phi_D)$ from Eq.~\ref{eq:generator_loss} \\
    $l_D= \mathcal{L}_D(T_{i}^{\text{query; generated}}, T_{i}^{\text{query}},  S^{\text{query}}_i; \phi_G, \phi_D)$ from Eq.~\ref{eq:discriminator_loss}\\
    update rolling sums $l_G^{\text{batch}} += l_G$ and $l_D^{\text{batch}} += l_D$\\
  }
  backpropagate $2^{\text{nd}}$ order updates $\nabla_{w_G} l_G^{\text{batch}}$ and $\nabla_{w_D}l_D^{\text{batch}}$ to $w_G$ and $w_D$
 }
 return good initializations $w_G$ and $w_D$ for both generator and discriminator.
 \caption{One Epoch MAML-Train Loop for U-Net Generator with Adversarial Loss.}
 \label{algo:maml_gan}
\end{algorithm}

\subsection{Experimental Details}
We run experiments using a single 32GB Nvidia Volta V100 GPU. For MAML optimization \citep{Arnold2020-ss} we use meta-batch sizes of 2, 3 or 4 events. For the corresponding joint training baselines we used $N_{\text{support}} + N_{\text{query}}$ frames from each event and comparable number of events to keep comparisons fair. We randomly split all SEVIR events into 9169 train, 1162 validation, and 1148 test tasks.  Joint training baselines and MAML outer loop optimizations are both performed using the Adam optimizer \citep{kingma2017adam} with learning rate $0.0002$ and momentum $0.5$. 

We resize input modalities to all have $192 \times 192$ resolution and keep the target at $384 \times 384$. The generator's encoder has four convolutional blocks, and the decoder has five. All generator blocks except for the last decoder layer use ReLU activation functions. The very last layer uses linear activation functions  to support z-score normalization for all four image modalities. 

\subsection{Results}

We test our multi-task few-shot formulation and demonstrate MAML provides empirical gains by comparing the performance of models trained using either meta-learning algorithms or joint training for both reconstruction and adversarial loss objectives. We find throughout all our experiments that meta-learning minimizes the reconstruction error compared to joint training. On the other hand, achieving better performance on the training objective does not always translate to higher weather evaluation metrics.

\subsubsection{Reconstruction Loss}
\begin{figure*}[t!]
    \centering
    \begin{subfigure}[t]{0.42\textwidth}
        \includegraphics[width=0.3\linewidth]{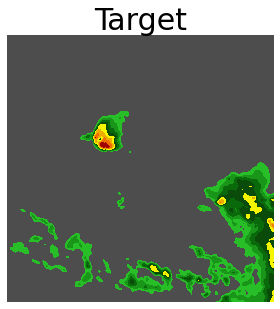}
        \includegraphics[width=0.3\linewidth]{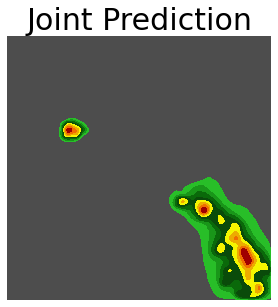}
        \includegraphics[width=0.3\linewidth]{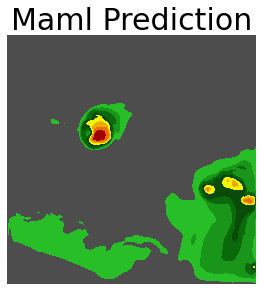}
    \caption{Target VIL test-frame and generated samples. Task-adaptation helps recognize sparse VIL regions.}
    \end{subfigure}\hfill
    \begin{subfigure}[t]{0.55\textwidth}
        \includegraphics[width=0.45\linewidth]{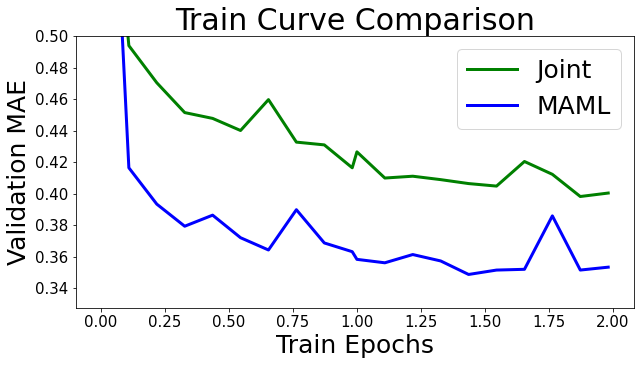}
        \includegraphics[width=0.47\linewidth]{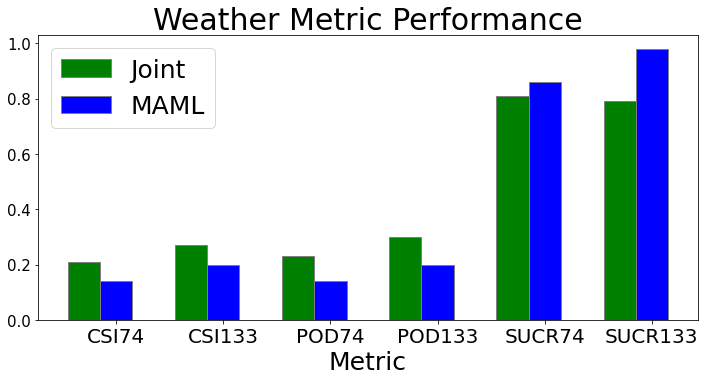}
    \caption{Validation mean absolute error throughout training and test-set evaluation of weather-specific metrics. MAML optimization leads to better train objective and SUCR but lower CSI and POD. }
    \end{subfigure}
    \caption{{\bf Reconstruction Loss Results: } Examples of generated samples; objective evolution through training; final evaluation on domain-specific metrics. \label{fig:reconstruction}}
\end{figure*}

We compare joint training with MAML that uses a single adaptation step for each event, evaluate model performance using weather metrics with two different thresholds (74 and 133), and summarize our results in Figure~\ref{fig:reconstruction}. We find that even though U-Nets trained with MAML achieve better performance on the optimization objective, these improvements do not consistently translate to gains on weather-specific evaluation. In particular, we see that finetuning to specific tasks leads to better precision but worse recall and IOU.  Limitations given by training with reconstruction loss, such as blurry outputs, remain. The task adaptation mechanism helps in this case to recognize that there are some storm events in the lower-left corner, although it is not very effective at predicting the correct shape of these low-intensity precipitations on a fine-grained scale.

\subsubsection{Adversarial Loss}

\begin{figure}[t]
    \centering    \includegraphics[width=0.5\linewidth]{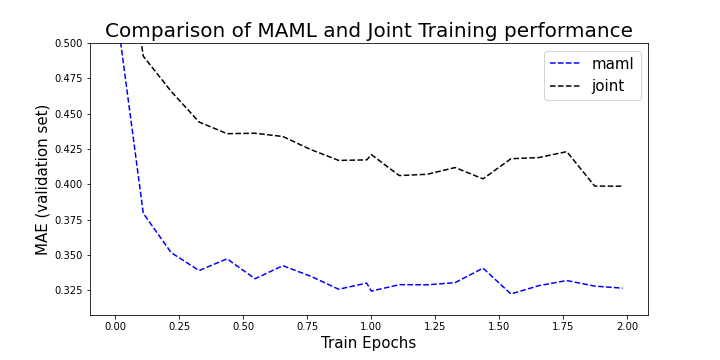}
    \caption{{\bf Adversarial loss - train curve.} MAML outperforms Joint Training. Evaluation is done on validation set throughout training. }
    \label{fig:gan_maml_vs_joint}
\end{figure}

We train generative adversarial networks using the second-order MAML procedure (on both the generator and discriminator networks,  as described in Algorithm~\ref{algo:maml_gan}) and the joint training baseline. We compare the evolution of the reconstruction error throughout training in Figure~\ref{fig:gan_maml_vs_joint} and notice MAML significantly helps in minimizing the training objective. We used $\lambda=10^2$ and $\eta=10^{-4}$ for this MAML curve. 

Next, we evaluate on meteorological metrics for all values of $\lambda$ and $\eta$, and summarize our results in Table~\ref{table:joint_gan} and \ref{table:maml_gan} for joint and MAML training, respectively. For joint adversarial training,  especially when evaluating with lower thresholds, we see the critical success index is fairly constant as we vary $\lambda$ while increasing $\lambda$ leads to lower recall and higher precision. This seems to suggest that placing more weight on the reconstruction loss will lead to predicting fewer high-valued VIL pixels. 

\begin{table}[t]
\caption{{\bf Adversarial Joint - evaluation.}  Test-set evaluation on meteorological metrics. }
\label{table:joint_gan}
\begin{center}
\begin{tabular}{r|r|r|r||r|r|r}
threshold&  \multicolumn{3}{c||}{74} &  \multicolumn{3}{c}{133} \\\cline{1-7}
 metric &  CSI  &  POD &  SUCR & CSI & POD & SUCR   \\\hline
$\lambda=10^2$ &  0.29 & 0.50  & 0.56  & 0.27 &  0.30 & 0.76          \\\hline
$\lambda=10^3$ &  0.29 &  0.46 & 0.58  &  0.29  &  0.35 &   0.71    \\\hline
$\lambda=10^4$ & 0.29 & 0.43  &  0.64 &  0.29  & 0.33  & 0.73  
\end{tabular}
\end{center}
\end{table}

\begin{figure}[t]
    \centering    \includegraphics[width=1\linewidth]{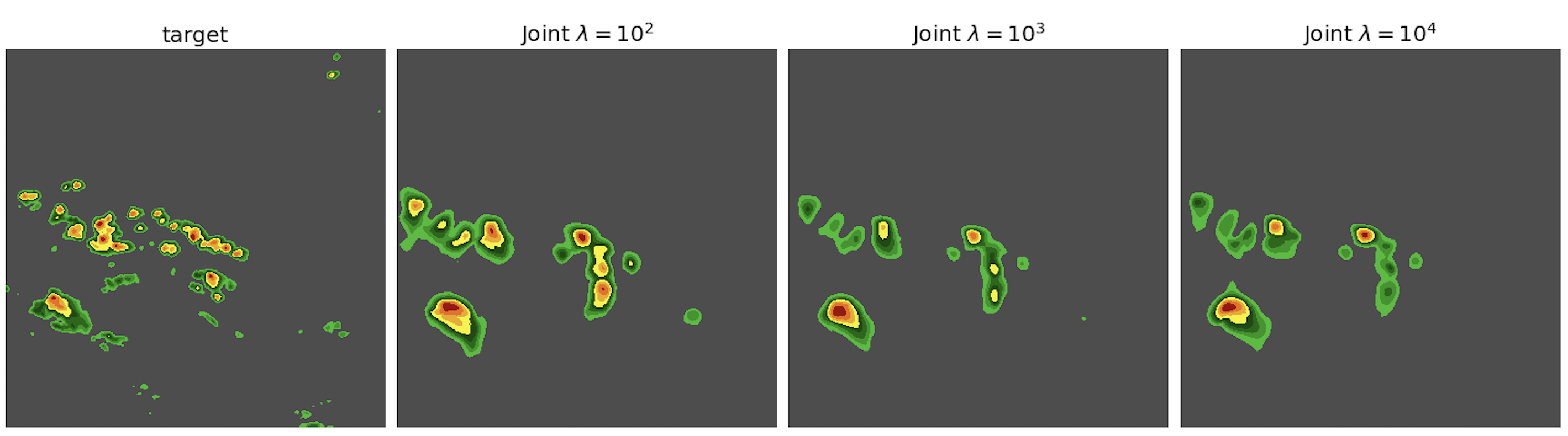}
    \caption{{\bf Adversarial Joint - generated samples.} Reconstruction loss biases the model towards sparser predictions.   }
    \label{fig:adversarial-joint-samples}
\end{figure}

For MAML adversarial training we do not identify any clear trend between hyperparameters $\lambda$ and $\eta$ and the values of meteorological metrics on the test-split. We believe this shows training is more unstable in this regime: instabilities are further exacerbated when optimizing a bilevel Nash equilibrium problem with gradient descent, as we did above. A comparison between Tables~\ref{table:joint_gan} and \ref{table:maml_gan} shows that, similarly to the case of reconstruction loss, MAML optimization leads to higher precision and lower recall.  After visually inspecting the generated samples we find that some of the models seem to exhibit mode collapse where the generated samples are not even realistic, while some of them do resemble the ground-truth. We present examples of samples successfully generated by models trained with MAML on adversarial loss below and note there is a large variance in the fraction of realistic samples across different models. This is not reflected in any of the evaluation metrics: we believe this further underscores that in image generation the correlation between good evaluation performance and high sample quality is rather weak.

\begin{table}[t]
\caption{{\bf Adversarial MAML - evaluation.}  Test-set evaluation on meteorological metrics. MAML models have higher precision and lower recall and IOU. }
\label{table:maml_gan}
\begin{center}
\begin{tabular}{r|r||r|r|r||r|r|r}
\multicolumn{2}{c||}{threshold}&  \multicolumn{3}{c||}{74} &  \multicolumn{3}{c}{133} \\\hline
\multicolumn{2}{c||}{metric}&   CSI  &  POD &  SUCR & CSI & POD & SUCR    \\\hline
\multirow{3}{*}{$\eta=10^{-4}$}& $\lambda=10^2$ & 0.14 & 0.16 & 0.93 & 0.24 & 0.26 & 0.90          \\\cline{2-8} 
& $\lambda=10^3$& 0.09  & 0.09 & 0.98 & 0.20 & 0.20 & 0.99
   \\\cline{2-8} 
& $\lambda=10^4$ & 0.13 & 0.21 & 0.91 & 0.21 & 0.32 & 0.87  \\\hline
\multirow{3}{*}{$\eta=10^{-5}$}& $\lambda=10^2$ & 0.19 & 0.23 & 0.87 & 0.23 & 0.27 & 0.84            \\\cline{2-8} 
& $\lambda=10^3$ & 0.17 & 0.20 & 0.90 & 0.25 & 0.29 & 0.87    \\\cline{2-8} 
& $\lambda=10^4$ & 0.12 & 0.15 & 0.93 & 0.22 & 0.26 & 0.91  
\end{tabular}
\end{center}
\end{table}

\begin{figure}[t]
    \centering    \includegraphics[width=1\linewidth]{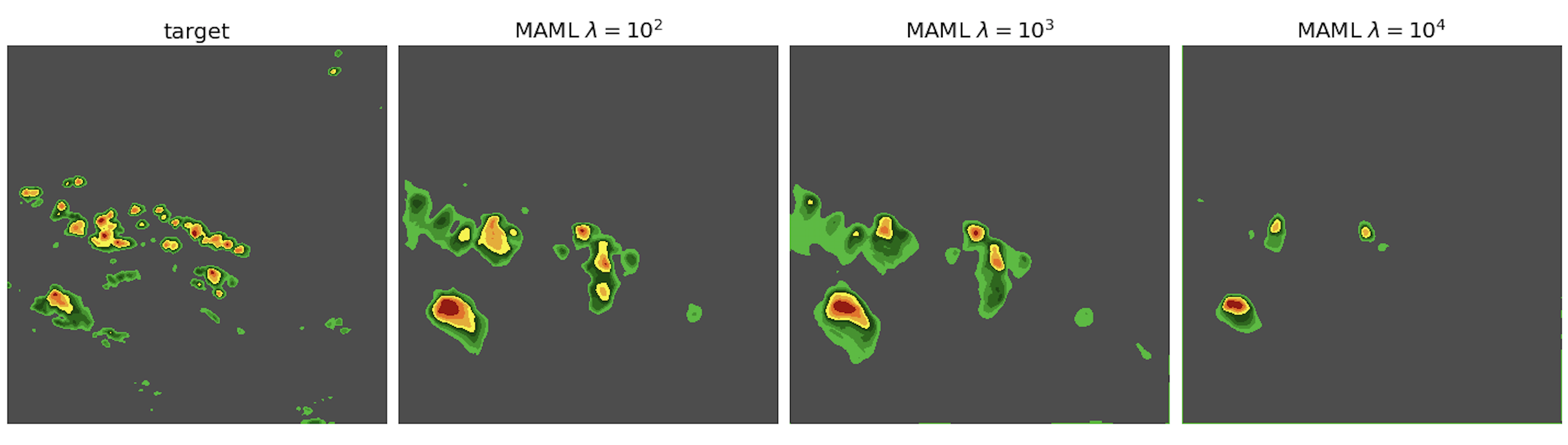}
   \caption{{\bf Adversarial MAML - generated samples.} Finetuning helps identify low-intensity VIL regions. }
    \label{fig:adversarial-maml-samples}
\end{figure}

Figures~\ref{fig:adversarial-joint-samples} and~\ref{fig:adversarial-maml-samples} compare samples generated by models trained on adversarial loss through either joint or MAML-based procedures for different values of $\lambda$. The MAML models all used an inner SGD learning rate of $10^{-5}$. We see that in this case the intuitions from the reconstruction loss setting are still valid and the task-adaptation inherent to MAML enables it to correctly generate low-intensity VIL data that joint-setting misses out on. We also confirm the aforementioned trend of higher $\lambda$ values leading to lower VIL values.

\section{Self-supervised Pre-training}

\subsection{Method}
We follow recent work in self-supervised pretraining which applies contrastive learning to convolutional networks before finetuning on classification tasks and improves downstream performance and data efficiency. We ask if these improvements extrapolate to our image-to-image setup. The main distinction between our scenario and those in previous work is that we can initialize only a fraction of our parameters through contrastive pretraining. 

We restrict our attention to the U-Net encoder parameters during the pretraining stage and follow the same network architecture as in Section~\ref{sec:fsbench}. Our experiments are inspired by the large-scale study on unsupervised spatiotemporal representation learning, conducted by~\citet{feichtenhofer2021large}. In particular, we focus on MoCoV3~\citep{chen2021empirical}, which is a state-of-the-art contrastive learning method, because~\citet{feichtenhofer2021large} identify the momentum contrast (MoCo) contrastive learning method as the most useful for spatiotemporal data.

\paragraph{Pre-training objective.} For a given representation $q$ of a query frame from the dataset, a positive key representation $k^+$ and a negative key representation $k^-$, the loss function increases the similarity between the representations within the positive pair $(q, k^+)$ and decreases the similarity within the negative pair $(q, k^-)$ respectively. All representations are normalized on the unit sphere and the similarity is the dot product (i.e. the cosine similarity, because the representations are normalized).

The loss is the InfoNCE loss~\citep{oord2018representation}, given as follows:
\begin{equation}
\hat{\mathcal{L}}_q = - \log \frac{\exp(p(q) \cdot \text{stropgrad}({k^+})+ / \tau)}{\exp(p(q) \cdot \text{stopgrad}({k^+}) / \tau) + \sum_{k^-} \exp(p(q) \cdot \text{stopgrad}({k^-}) / \tau)}
\end{equation}
for a temperature parameter $\tau$ and a predictor MLP $p$, which is a  two layer MLP, with input dimension 128, hidden dimension 2048, output dimension 128, BatchNorm and ReLU in the hidden layer activation. Following~\citep{chen2021empirical}, the gradients are not backpropagated through $k^{\{+,-\}}$ and the encoder representations both for keys and queries are obtained after a composition of the backbone and the projector (which is a two layer MLP, with dimensions [256, 2048, 128] with BatchNorm and ReLUs in between the hidden layers, and ending with a BatchNorm with no trainable affine parameters). Additionally, the branch for key representations follows the momentum update policy $\theta_\text{k} \leftarrow m\theta_\text{k} + (1 - m) \theta_\text{q}$ from~\citep{he2020momentum} with momentum parameter $m=0.999$, where $\theta_\text{k}$ are the weights in the key branch and $\theta_\text{q}$ are the weights in the query branch.

{\bf Data Augmentations.} Another difficulty particular to our setup is the problem of choosing data augmentations the input domain is invariant to because weather modalities have different invariances than natural images: for example, the popular color jitter transformation is not applicable here, because image-to-image translation is sensitive to color. From the standard augmentations, the ones we consider are: random resized crops, random horizontal flips, gaussian noise, gaussian blur, random vertical flips and random rotation. We also further exploit the temporal structure of SEVIR to obtain ``natural'' augmentations, which we introduce next.

\begin{figure}[t]
    \centering    \includegraphics[width=1\linewidth]{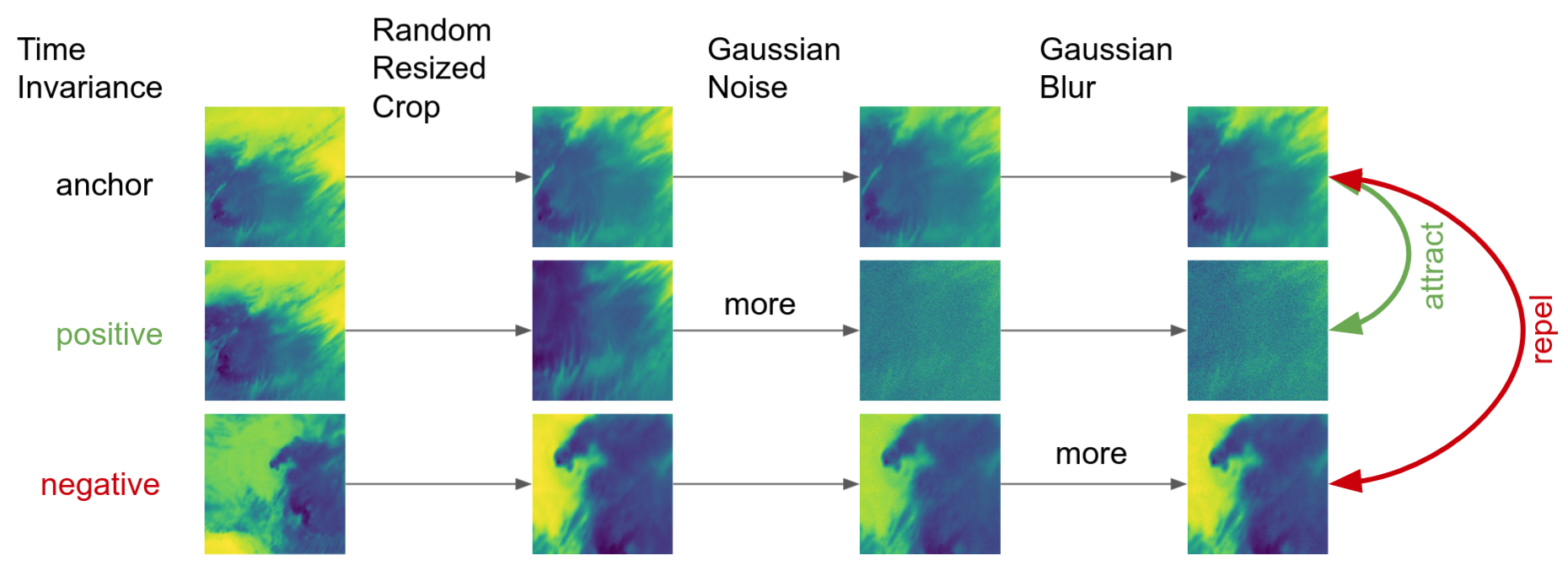}
    \caption{{\bf Augmentations for the contrastive learning experiment} By indicating ``more'' we show examples of a larger magnitude of the augmentation being applied.}
    \label{fig:augmentations}
\end{figure}

{\bf Natural augmentations.} We further consider using the temporal structure of SEVIR for augmentations, as follows. Each event consists of 49 frames, so we anchor every even frame as query frame and use every odd frame as key frame.
For each query frame, to obtain $q$ and $k^{+}$ we apply the following stochastic transformations to the frame twice: random resized crops using scale (0.8, 1.0); random horizontal flips with probability 0.5, pixel-wise gaussian noise sampled from the normal distribution $\mathcal{N}(0, 0.1)$ with probability 0.5, gaussian blur with kernel size 19, random vertical flips with probability 0.2, random rotation by angle unformly chosen in $(-\pi/6, \pi/6)$.
The rest of the augmentation arguments follow the default in the Torchvision library\footnote{\url{https://pytorch.org/vision/stable/transforms.html}}. In Figure~\ref{fig:augmentations} we present a conceptual visualization of the transforms. To obtain $k^-$ we apply the above stochastic transformations to the corresponding key frame once.


{\bf Training hyperparameters.} Our experiments use the following architectural choices: mini-batch, consisting of 3 events with 24 frames for queries and key 24 frames for keys each; 0.015 base learning rate for the baseline SimCLR and SimSiam and; 100 pre-training epochs; standard cosine decayed learning rate; 5 epochs for the linear warmup; 0.0005 weight decay value; SGD with momentum 0.9 optimizer. We report the joint training reconstruction loss experimental setup by finetuning the checkpoint obtained from pretraining.
All parameters are in a Pytorch-like style.

\begin{figure}[t]
    \centering    \includegraphics[width=1\linewidth]{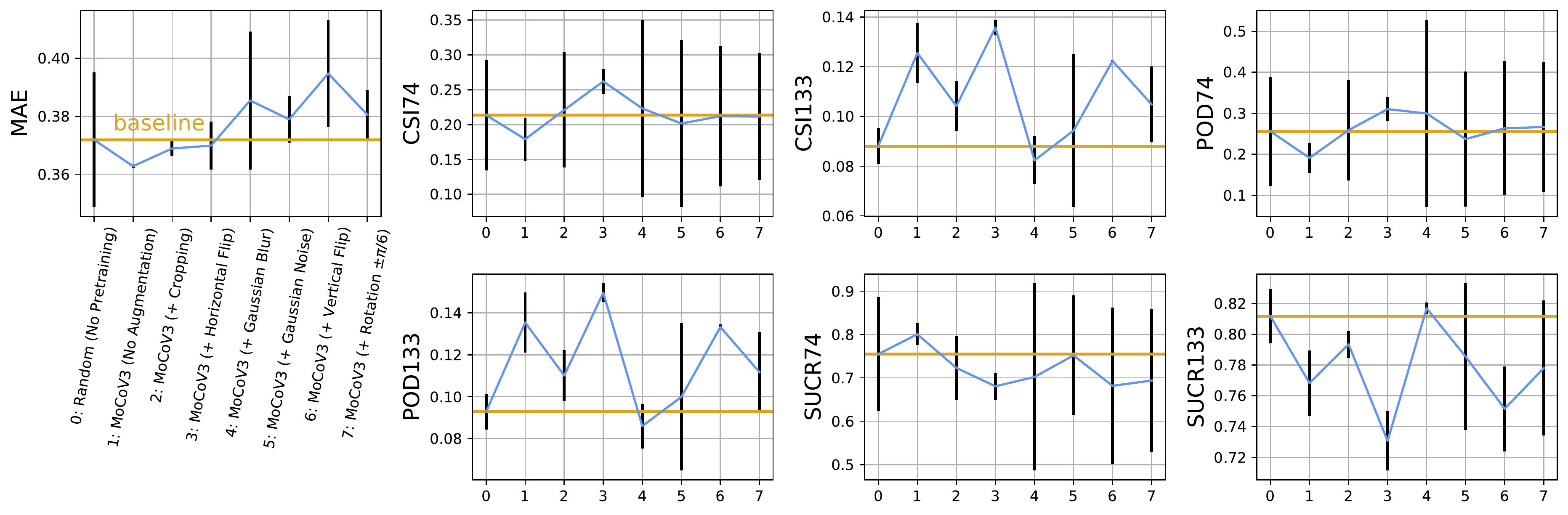}
   \caption{{\bf Contrastive Learning for SEVIR.} For mean absolute error lower is better. For every other evaluation measure, higher is better.}
    \label{fig:cl_sevir}
\end{figure}

\subsection{Results}
\begin{figure}[t]
    \centering    \includegraphics[width=1\linewidth]{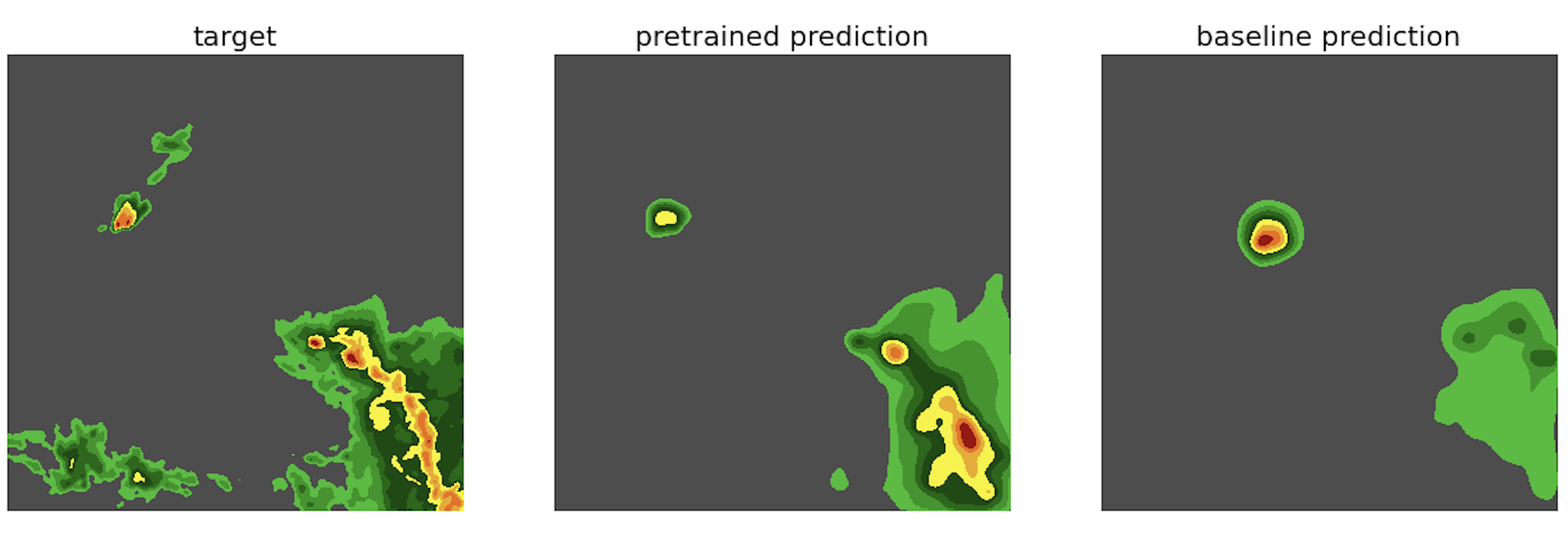}
    \caption{{\bf Pretrained encoder - generated samples} Pretrained models better identify the sparse high VIL values. }
    \label{fig:pretrained}
\end{figure}

In Figure~\ref{fig:cl_sevir} we report our results. Firstly, for mean absolute error we find marginal yet somewhat consistent gains up to level 3 augmentation. Secondly, we also evaluate on meteorological metrics. We find that even though pretraining has a marginal effect on the reconstruction loss train objective, it often provides important gains on domain-specific evaluation criteria. We highlight the large improvement in CSI133 and POD133, which stems mostly from significant improvements in precision. We observe that up to level 4 MoCoV3 augmentation we obtain improvements throughout all measures with the contrastive pretraining. Finally, we show example samples in figure~\ref{fig:pretrained} and find that pretraining the U-Net encoder leads to better performance in high-VIL regions. 

\section{Conclusion and Future Work}

\subsection{Conclusion: novel few-shot multi-task image-to-image translation}
We formulated a novel few-shot multi-task image-to-image translation problem leveraging spatio-temporal structure in a large-scale storm event dataset. We provided several benchmarks for this problem and considered two optimization procedures (joint training and gradient-based meta-learning) and two loss functions (reconstruction and adversarial). We trained U-Nets in all these regimes and presented each model's performance, as well as evaluated on various domain-specific metrics. We discussed the advantages and disadvantages of each of these. In this process we also explored a training method unexplored until now to the best of our knowledge: meta-learning adversarial GANs with second-order gradient updates. Additionally, we explored pretraining U-Net encoder parameters using various augmentations in both the spatial and temporal domains. 


\subsection{Future work: improving performance and stability}
There are numerous tricks for training GANs that have been shown to work well in practice for natural image generation. An interesting research direction would be exploring if these gains extend to our meteorological domain. Two of these techniques are applying spectral normalization to the discriminator network and updating the generator network more often than the discriminator.

We have not fully explored the interplay between adversarial training and MAML's bilevel optimization, and we believe it would also be very interesting to further develop this aspect of our work. The most immediate next step could be meta-learning just a subset of the networks' parameters.

Another interesting direction would be applying importance sampling or even curriculum learning techniques to the training schedule. An important difference between SEVIR and the natural images datasets we are more accustomed to is that not all events are equally informative: our models can presumably learn much more from complex storms than from frames taken during calm weather where the VIL and lighting frames are very sparse, and the IR imagery has very little variance. 




\subsection*{Acknowledgements}

The authors acknowledge the MIT SuperCloud and Lincoln Laboratory Supercomputing Center \citep{reuther2018interactive} for providing HPC and consultation resources that have contributed to the research results reported within this paper/report.

Research was sponsored by the United States Air Force Research Laboratory and the United States Air Force Artificial Intelligence Accelerator and was accomplished under Cooperative Agreement Number FA8750-19-2-1000. The views and conclusions contained in this document are those of the authors and should not be interpreted as representing the official policies, either expressed or implied, of the United States Air Force or the U.S. Government. The U.S. Government is authorized to reproduce and distribute reprints for Government purposes notwithstanding any copyright notation herein.

This material is also based in part upon work supported by the Air Force Office of Scientific Research under the award number
FA9550-21-1-0317 and the U. S. Army Research Office through the Institute for Soldier Nanotechnologies at MIT, under Collaborative Agreement Number W911NF-18-2-0048. This work is also supported in part by the the National Science Foundation under Cooperative Agreement PHY-2019786
(The NSF AI Institute for Artificial Intelligence and Fundamental Interactions, \url{http://iaifi.org/}).

\bibliography{iclr2020_conference}
\bibliographystyle{iclr2020_conference}


\end{document}